\documentclass[11pt,a4paper]{article}
\usepackage[hyperref]{acl2021}
\usepackage{times}
\usepackage{xcolor}
\usepackage{latexsym}

\usepackage{CJKutf8}
\newcommand{\cn}[1]{{\begin{CJK*}{UTF8}{gbsn}#1\end{CJK*}}}

\usepackage[whole]{bxcjkjatype} 

\usepackage{stfloats}
\usepackage{float}
\usepackage{afterpage}
\usepackage{xcolor}

\usepackage{amsmath,amssymb,mathtools}
\usepackage{pifont}

\usepackage[normalem]{ulem}

\makeatletter
\def\new@fontshape{}
\makeatother
\usepackage{gb4e}
\noautomath

\makeatletter
\renewcommand{\@subex}[2]{%
    \settowidth{\labelwidth}{#1}\itemindent\z@\labelsep#2%
    \ifnum\the\@xnumdepth=1%
        \topsep 7\p@ plus2\p@ minus3\p@\itemsep3\p@ plus2\p@\else%
        \topsep1.5\p@ plus\p@\itemsep1.5\p@ plus\p@\fi%
    \parsep\p@ plus.5\p@ minus.5\p@%
    \leftmargin\labelwidth%
    \ifnum\the\@xnumdepth=2%
    \else\advance\leftmargin#2\relax\fi
}
\makeatother

\usepackage{hyperref}
\usepackage[noabbrev, nameinlink, capitalize]{cleveref}

\renewcommand{\ref}{\cref}

\crefname{section}{第}{第}
\creflabelformat{section}{#2#1節#3}
\crefname{figure}{図}{図}
\crefname{appendix}{付録}{付録}
\crefname{footnote}{脚注}{脚注}

\usepackage{listings,jvlisting}
\usepackage{booktabs}
\usepackage{multirow}
\usepackage{tabularx}

\usepackage{tikz}
\usepackage{tikz-qtree}
\usetikzlibrary{positioning}
\usetikzlibrary{shapes.misc}
\usetikzlibrary{arrows.meta}

\usepackage{microtype}

\aclfinalcopy 

\title{LLMs Struggle with NLI for Perfect Aspect: A Cross-Linguistic Study in Chinese and Japanese}

\author{
Jie Lu$^{1}$ \quad Du Jin$^{1}$ \quad Hitomi Yanaka$^{1,2}$ \\
$^{1}$the University of Tokyo \quad $^{2}$RIKEN \\
\texttt{\{lujie2001yoshino, dujin728\}@gmail.com}\\
\texttt{hyanaka@g.ecc.u-tokyo.ac.jp}
}

\date{}

\begin{document}
\maketitle
\begin{abstract}
Unlike English, which uses distinct forms (e.g., had, has, will have) to mark the perfect aspect across tenses, Chinese and Japanese lack separate grammatical forms for tense within the perfect aspect, which complicates Natural Language Inference (NLI).
Focusing on the perfect aspect in these languages, we construct a linguistically motivated, template-based NLI dataset (1,350 pairs per language). 
Experiments reveal that even advanced LLMs struggle with temporal inference, particularly in detecting subtle tense and reference-time shifts. 
These findings highlight model limitations and underscore the need for cross-linguistic evaluation in temporal semantics. Our dataset is available at \url{https://github.com/Lujie2001/CrossNLI}.
\end{abstract}

\section{Introduction}
\label{sec:intro}
Recent advances in large language models (LLMs) have raised important questions about the depth and limits of their language understanding.
While these models perform well on many standardized benchmarks, most such evaluations are heavily centered on English and often overlook linguistic features that are specific to other languages.

This paper focuses on whether LLMs have human-like understanding of the perfect aspect of punctual verbs in Chinese and Japanese.
Although both languages exhibit features that differ from English (See Section~\ref{sec:Perfect Aspect}), there has been no systematic investigation of how the perfect aspect is represented or interpreted in these languages within the NLI framework.

To address this gap, we construct a challenging dataset targeting the interpretation of the perfect aspect with punctual verbs (e.g., \textit{die}) in Chinese and Japanese. 
Our dataset is linguistically motivated, template-based, and contains 1,350 sentence pairs per language.

Our contributions are as follows: 
\begin{enumerate}
\item We construct a bilingual NLI dataset focused on perfect aspect in Chinese and Japanese.
\item Our analysis reveals that even the current state-of-the-art LLMs repeatedly fail on specific types of problems in our dataset, indicating that they have not fully acquired a robust or generalizable understanding of the perfect aspect in Chinese and Japanese.
\end{enumerate}

\section{Background}
\label{sec:bk}

\subsection{Perfect Aspect in Chinese and Japanese}
\label{sec:Perfect Aspect}
Following \citet{reichenbach1947}, we analyze the temporal interpretation of the perfect aspect by appealing to a three-way temporal distinction: Speech Time (S), Event Time (E), and Reference Time (R).
In Reichenbach's framework, different tenses can be interpreted as different relations between S, E, and R. 
In the past, R occurs before S; in the present, R and S are simultaneous; in the future, R is after S. 
Furthermore, in the perfect aspect, E always occurs before R, regardless of tense.
In Example (\ref{exp:Perfect Aspect}),  E (``Hanako graduates") precedes R (``Taro gets PhD"), thus the overall temporal relation of the sentence is (S $<$ E $<$ R). 
Here, A $<$ B signifies that A takes place before B.

\begin{exe}
    \ex \label{exp:Perfect Aspect} When Taro gets his PhD next year, Hanako will have graduated from college.
\end{exe}

In addition, the time interval between E and R is specified by adding temporal adverbs in the main clause  (e.g., ``When Taro gets his PhD next year, Hanako will have graduated from college \textit{3 months ago}").

In English, the perfect aspect is marked differently depending on tense (e.g., had, has, will have).
However, Chinese and Japanese do not morphologically vary aspect markers across tenses. 
Chinese typically uses the marker ``\textit{-le}(了)"\citep{klein2000,mochizuki1997} to indicate the perfect aspect regardless of tense and relies on temporal adverbs or context to convey temporal information. 
Japanese expresses the perfect aspect using the auxiliary ``\textit{-tei-}(-てい-)"\citep{kudo1995, iori2001}, combined with either the past ``\textit{-tei-ta}(-てい-た)" or non-past ``\textit{-tei-ru}(-てい-る)" form, reflecting its binary tense system.\footnote{Other markers such as ``guo" (Chinese), ``zhe" (Chinese), and ``-ta" (Japanese) may express perfect meanings; however, this paper primarily focuses on the prototypical ``-le" and ``-tei-".}

These aspect markers are also used in other contexts and are not exclusively used to express the perfect aspect. 
For example, Chinese ``\textit{le}" may also serve as a modal particle to express urgency or emotional emphasis (e.g., ``太好了!" means ``great!"). 
Because such non-perfect uses dominate everyday usage, we hypothesized that LLMs may struggle to generalize the meaning of the perfect aspect in these languages.


\subsection{Temporal NLI Datasets}
\label{sec:Previous studies}
There are already some NLI datasets that focus on aspect~\citep{kober2019,prus2024}.
\citet{kober2019} introduced a carefully curated NLI dataset with a specific focus on tense and aspect.
However, these studies focus only on English.

Several studies~\cite{ocnli,yanaka-mineshima-2021-assessing,yanaka-mineshima-2022-compositional,sugimoto2024} have addressed NLI tasks involving challenging linguistic phenomena in Japanese and Chinese, but they rarely involve NLI tasks focusing on the perfect aspect.
OCNLI \cite{ocnli} is a Chinese NLI dataset, and JaNLI \cite{yanaka-mineshima-2021-assessing} and JSICK \cite{yanaka-mineshima-2022-compositional} are Japanese NLI datasets. 
However, they scarcely address temporal inference. 
Jamp\_sp \cite{sugimoto2024} is a Japanese temporal inference dataset, but it does not systematically investigate inference tasks concerning the perfect aspect.


\section{Dataset}
\label{sec:dataset}
Based on tense (past (Pst), present (Pres), future (Fut)) and the presence (t) or absence (None) of a temporal adverb in the main clause discussed in Section~\ref{sec:Perfect Aspect}, 
we designed six Japanese sentence templates based on linguistic literature~\citep{kudo1995} and created corresponding Chinese templates.
By using these sentence templates as premises and hypotheses, we constructed 30 premise--hypothesis pairs $(P,H)$ of NLI problems for Japanese and Chinese, respectively.
Since the perfect aspect with punctual verbs expresses a stable temporal relation in sentences, each $(P,H)$ pair is theoretically expected to have a unique correct label (\textit{entailment} or \textit{non-entailment}) under various punctual verb phrases (See a and b in Example (\ref{exp:reviewer})).
This enables us to generate a large number of $(P,H)$ pairs with entailment labels by inserting different lexical items semi-automatically.

\begin{exe}
    \ex\label{exp:reviewer}
        \begin{xlist}
            \ex Pres(t): Hanako has already \textbf{been dead} for 3 months. \\
    $\Rightarrow$ Pres: Hanako has already \textbf{been dead}.
            \ex Pres(t): Hanako has already \textbf{graduated from college} for 3 months. \\
    $\Rightarrow$ Pres: Hanako has already \textbf{graduated from college}.
        \end{xlist} 
\end{exe}

The examples of sentence templates with labels for Chinese are shown in Table~\ref{tab:Example of NLI tasks}.
Full examples of $(P,H)$ pairs (Table~\ref{tab:question_example}) and sentence templates in Chinese and Japanese (Tables~\ref{tab:dataset_example_cn} and Table~\ref{tab:dataset_example_jp}) can be found in Appendix~\ref{sec:templates}.

\begin{table*}[t]
\footnotesize
\centering
\renewcommand{\arraystretch}{1.15}
\begin{tabular}{@{}p{2.3cm}p{11.7cm}@{}}
\toprule
\textbf{Categories} & \textbf{Template Example} \\
\midrule
P: Pst(t)  
& \textcolor{red}{\texttt{\texttt{[Event-Past]}}} \textit{的\cn{时候},} \textcolor{blue}{\texttt{[NP]}} \textit{\cn{已经}} \textcolor{orange}{\texttt{[VP]}} \textcolor{teal}{\texttt{[TIME]}} \textbf{了}. \\
(E $<$ R $<$ S) & \textcolor{red}{太郎去年取得博士学位} \textit{的\cn{时候},} \textcolor{blue}{花子} \textit{\cn{已经}} \textcolor{orange}{\cn{死}} \textcolor{teal}{三个月} \textbf{了}. \\
& ``When \textcolor{red}{Taro got his PhD last year}, \textcolor{blue}{Hanako} \textit{had} already \textcolor{orange}{been dead} \textcolor{teal}{for 3 months}." \\

$\Rightarrow$ H\textsubscript{1}: Pst  
& \textcolor{red}{\texttt{\texttt{[Event-Past]}}} \textit{的\cn{时候},} \textcolor{blue}{\texttt{[NP]}} \textit{\cn{已经}} \textcolor{orange}{\texttt{[VP]}} \textbf{了}. \\
(E $<$ R $<$ S) & \textcolor{red}{太郎去年取得博士学位} \textit{的\cn{时候},} \textcolor{blue}{花子} \textit{\cn{已经}} \textcolor{orange}{\cn{死}} \textbf{了}. \\
& ``When \textcolor{red}{Taro got his PhD last year}, \textcolor{blue}{Hanako} \textit{had} already \textcolor{orange}{been dead}." \\

$\not\Rightarrow$ H\textsubscript{2}: Pres(t)  
& \textcolor{blue}{\texttt{[NP]}} \textit{\cn{已经}} \textcolor{orange}{\texttt{[VP]}} \textcolor{teal}{\texttt{[TIME]}} \textbf{了}. \\
(E $<$ S $=$ R) & \textcolor{blue}{花子} \textit{\cn{已经}} \textcolor{orange}{\cn{死}} \textcolor{teal}{三个月} \textbf{了}. \\
& ``\textcolor{blue}{Hanako} \textit{has} already \textcolor{orange}{been dead} \textcolor{teal}{for 3 months}." \\

$\Rightarrow$ H\textsubscript{3}: Pres  
& \textcolor{blue}{\texttt{[NP]}} \textit{\cn{已经}} \textcolor{orange}{\texttt{[VP]}} \textbf{了}. \\
(E $<$ S $=$ R) & \textcolor{blue}{花子} \textit{\cn{已经}} \textcolor{orange}{\cn{死}} \textbf{了}. \\
& ``\textcolor{blue}{Hanako} \textit{has} already \textcolor{orange}{been dead}." \\

$\not\Rightarrow$ H\textsubscript{4}: Fut(t) 
& \textcolor{red}{\texttt{[Event-Future]}} \textit{的\cn{时候},} \textcolor{blue}{\texttt{[NP]}} \textit{\cn{已经}} \textcolor{orange}{\texttt{[VP]}} \textcolor{teal}{\texttt{[TIME]}} \textbf{了}. \\
(S $<$ E $<$ R) & \textcolor{red}{太郎明年取得博士学位} \textit{的\cn{时候},} \textcolor{blue}{花子} \textit{\cn{已经}} \textcolor{orange}{\cn{死}} \textcolor{teal}{三个月} \textbf{了}. \\
& ``When \textcolor{red}{Taro gets his PhD next year}, \textcolor{blue}{Hanako} \textit{will have} already \textcolor{orange}{been dead} \textcolor{teal}{for 3 months}." \\

$\Rightarrow$ H\textsubscript{5}: Fut  
& \textcolor{red}{\texttt{[Event-Future]}} \textit{的\cn{时候},} \textcolor{blue}{\texttt{[NP]}} \textit{\cn{已经}} \textcolor{orange}{\texttt{[VP]}} \textbf{了}. \\
(S $<$ E $<$ R) & \textcolor{red}{太郎明年取得博士学位} \textit{的\cn{时候},} \textcolor{blue}{花子} \textit{\cn{已经}} \textcolor{orange}{\cn{死}} \textbf{了}. \\
& ``When \textcolor{red}{Taro gets his PhD next year}, \textcolor{blue}{Hanako} \textit{will have} already \textcolor{orange}{been dead}." \\
\bottomrule
\end{tabular}
\caption{Template examples of premise and hypothesis sentences in Chinese. 
In category column, the symbol (t) indicates the presence of a temporal adverb in the main clause.
The slot \texttt{[Event-Past]} and \texttt{[Event-Future]} is a subordinate clause containing a temporal expression referring to the past or future, such as ``太郎去年取得博士学位" (``Taro got his PhD last year").
$\Rightarrow$ indicates \textit{entailment} and $\not\Rightarrow$ indicates \textit{non-entailment}.}
\label{tab:Example of NLI tasks}
\end{table*}



We manually collected 45 sets of common lexical items (nouns and punctual verbs) and clauses to fill our templates.
To minimize semantic influence, the items were designed to maintain one-to-one semantic correspondence between Chinese and Japanese.
In total, we generated 1,350 $(P,H)$ pairs for each language, comprising 405 instances labeled as entailment and 945 instances labeled as non-entailment.

Some studies have noted that uncertainty may arise in NLI tasks when temporality is involved~\citep{kober2019,Pavlick2019}.
To address this issue, we limited the verb types to punctual verbs that denote irreversible changes (e.g., \textit{die}).

To validate the reliability of the sentences, all instances in the dataset underwent rigorous review and were refined by native speakers. 
Additionally, to ensure labeling reliability, multiple native speakers independently annotated 30 different $(P,H)$ pairs. 
Under a majority voting scheme, their judgments consistently matched the gold labels, demonstrating high inter-annotator agreement.\footnote{We collected answers from seven native Chinese speakers and three native Japanese speakers. The average match rate between the Chinese responses and the golden label is 94\%, while Japanese is 100\%.}

\section{Experimental Setup}
\label{sec:setup}
We conducted experiments on multilingual LLMs and LLMs with enhanced monolingual capability with varying parameter scales. 
The multilingual models we used include GPT-4 (gpt-4-0613), Claude 3.5 (claude-3-5-sonnet-20241022), Deepseek-V3 (deepseek-chat), and Llama3.1\footnote{\tiny\url{hf.co/collections/meta-llama/llama-31-669fc079a0c406a149a5738f}} (8B and 70B). 
The LLMs with enhanced monolingual capability include the Chinese models Qwen3\footnote{\tiny \url{hf.co/collections/Qwen/qwen3-67dd247413f0e2e4f653967f}} (8B and 32B) and the Japanese models Swallow\footnote{\tiny \url{hf.co/collections/tokyotech-llm/gemma-2-swallow-67f2bdf95f03b9e278264241}} (9B and 27B).
These models cover both multilingual and language-specialized types.

Each model received every premise--hypothesis pair in the corresponding language, together with an instructional prompt that introduces the NLI task and asks whether the premise entails the hypothesis. 
Model predictions were then compared with gold labels to compute classification accuracy. 
All experiments were conducted in a zero-shot setting. 
Our Japanese prompts were adapted from \cite{sugimoto2024} and then translated into Chinese by native speakers.
The full Chinese and Japanese prompts are provided in Appendix \ref{sec:prompts}. 

\section{Results and Discussion}
\label{sec:result}
\begin{table*}[!ht]
\scriptsize
\setlength{\tabcolsep}{2.5pt}
\renewcommand{\arraystretch}{0.9}
\centering
\begin{tabular}{@{}lcccccccccc@{}}
\hline
\textbf{Tense of $(P,H)$} & \textbf{Label} & \textbf{\scriptsize GPT-4} & \textbf{\scriptsize Claude3.5} & \textbf{\scriptsize Deepseek-v3} & \textbf{\scriptsize Llama-8B} & \textbf{\scriptsize Llama-70B} & \textbf{\scriptsize Qwen3-8B} & \textbf{\scriptsize Qwen3-32B} & \textbf{\scriptsize Swallow-9B} & \textbf{\scriptsize Swallow-27B} \\
\hline
\textbf{(Pst(t), Pres(t))} & \textbf{N} & \textbf{0.0/0.0} & \textbf{77.8/2.2} & \textbf{0.0/0.0} & \textbf{0.0/17.8} & \textbf{0.0/0.0} & \textbf{0.0/0.0} & \textbf{0.0/0.0} & \textbf{0.0/0.0} & \textbf{0.0/0.0} \\
(Pst(t), Pres)    & \textbf{E} & 100.0/100.0 & 100.0/97.8 & 100.0/100.0 & 100.0/57.8 & 100.0/95.6 & 100.0/100.0 & 100.0/100.0 & 95.6/62.2 & 100.0/62.2 \\
(Pst, Pres(t))    & N & 100.0/100.0 & 100.0/100.0 & 100.0/100.0 & 0.0/100.0 & 93.3/80.0 & 91.1/62.2 & 51.1/75.6 & 93.3/62.2 & 15.6/62.2 \\
\textbf{(Pst, Pres)}       & \textbf{E} & \textbf{100.0/100.0} & \textbf{95.6/84.4} & \textbf{100.0/100.0} & \textbf{100.0/62.2} & \textbf{100.0/88.9} & \textbf{100.0/100.0} & \textbf{100.0/100.0} & \textbf{100.0/60.0} & \textbf{100.0/60.0} \\
\textbf{(Fut(t), Pres(t))} & \textbf{N} & \textbf{6.7/0.0} & \textbf{91.1/24.4} & \textbf{0.0/0.0} & \textbf{0.0/51.1} & \textbf{8.9/8.9} & \textbf{0.0/0.0} & \textbf{0.0/2.2} & \textbf{0.0/0.0} & \textbf{0.0/0.0} \\
\textbf{(Fut(t), Pres)}    & \textbf{N} & \textbf{0.0/0.0} & \textbf{53.3/20.0} & \textbf{0.0/0.0} & \textbf{2.2/62.2} & \textbf{0.0/37.8} & \textbf{0.0/0.0} & \textbf{2.2/2.2} & \textbf{13.3/4.4} & \textbf{6.7/8.9} \\
(Fut, Pres(t))    & N & 100.0/97.8 & 100.0/100.0 & 100.0/100.0 & 11.1/95.6 & 97.8/93.3 & 97.8/46.7 & 48.9/82.2 & 97.8/62.2 & 51.1/60.0 \\
\textbf{(Fut, Pres)} & \textbf{N} & \textbf{0.0/0.0} & \textbf{86.7/42.2} & \textbf{2.2/0.0} & \textbf{0.0/51.1} & \textbf{0.0/73.3} & \textbf{0.0/0.0} & \textbf{0.0/33.3} & \textbf{0.0/24.4} & \textbf{2.2/15.6} \\
\hline
\end{tabular}
\caption{Model accuracy (\%) when the premise is in the past or future, and the hypothesis is in the present tense. 
Left side of ``/" shows accuracy in Chinese cases, and the right side shows Japanese cases.
E indicates entailment labels and N indicates non-entailment labels.
The rows in boldface indicate the questions with lexical overlap.}
\label{tab:model_time_performance_clean}
\end{table*}
\begin{table}[ht]
\centering
\begin{tabular}{lcc}
\hline
\textbf{Model} & \textbf{Language} & \textbf{Accuracy (E / N)} \\
\hline
Llama-8B   & CN & \textbf{92.6\%} / \textbf{13.5\%} \\
          & JA & 45.2\% / 71.3\% \\
Qwen3-8B   & CN & 44.6\% / 74.6\% \\
           & JA & 62.7\% / 71.4\% \\
Swallow-9B & CN & 46.9\% / 80.2\% \\
           & JA & 32.6\% / 48.4\% \\
\hline
\end{tabular}
\caption{The differences in accuracy between \textit{entailment} and \textit{non-entailment} cases for Llama-8B, Qwen3-8B and Swallow-9B.}
\label{tab:Analyzing Llama8B}
\end{table}

Table \ref{tab:model_language_accuracy} shows the average accuracy of tested models on our dataset.
Figure \ref{fig: Results of GPT-4} shows the detailed results of GPT-4.
See Appendix \ref{app:detailed results of Chinese} for detailed results of other models.

\begin{table}[t]
\small
\centering
\begin{tabular}{lc}
\hline
\textbf{Model} & \textbf{Accuracy (CN / JA)} \\
\hline
GPT-4         & 80.6\% / 72.3\% \\
Claude3.5     & 91.5\% / 76.7\% \\
Deepseek-v3   & 77.3\% / 70.1\% \\
Llama-8B      & 37.3\% / 65.6\% \\
Llama-70B     & 75.8\% / 72.3\% \\
Qwen3-8B      & 74.2\% / 68.8\%    \\
Qwen3-32B     & 51.4\% / 56.6\%    \\
Swallow-9B    & 70.2\% / 43.6\%    \\
Swallow-27B   & 54.9\% / 42.7\%    \\
\hline
\end{tabular}
\caption{Overall accuracy of each model on our dataset.}
\label{tab:model_language_accuracy}
\end{table}

\begin{figure}[t]
    \centering
    \includegraphics[width=1\linewidth]{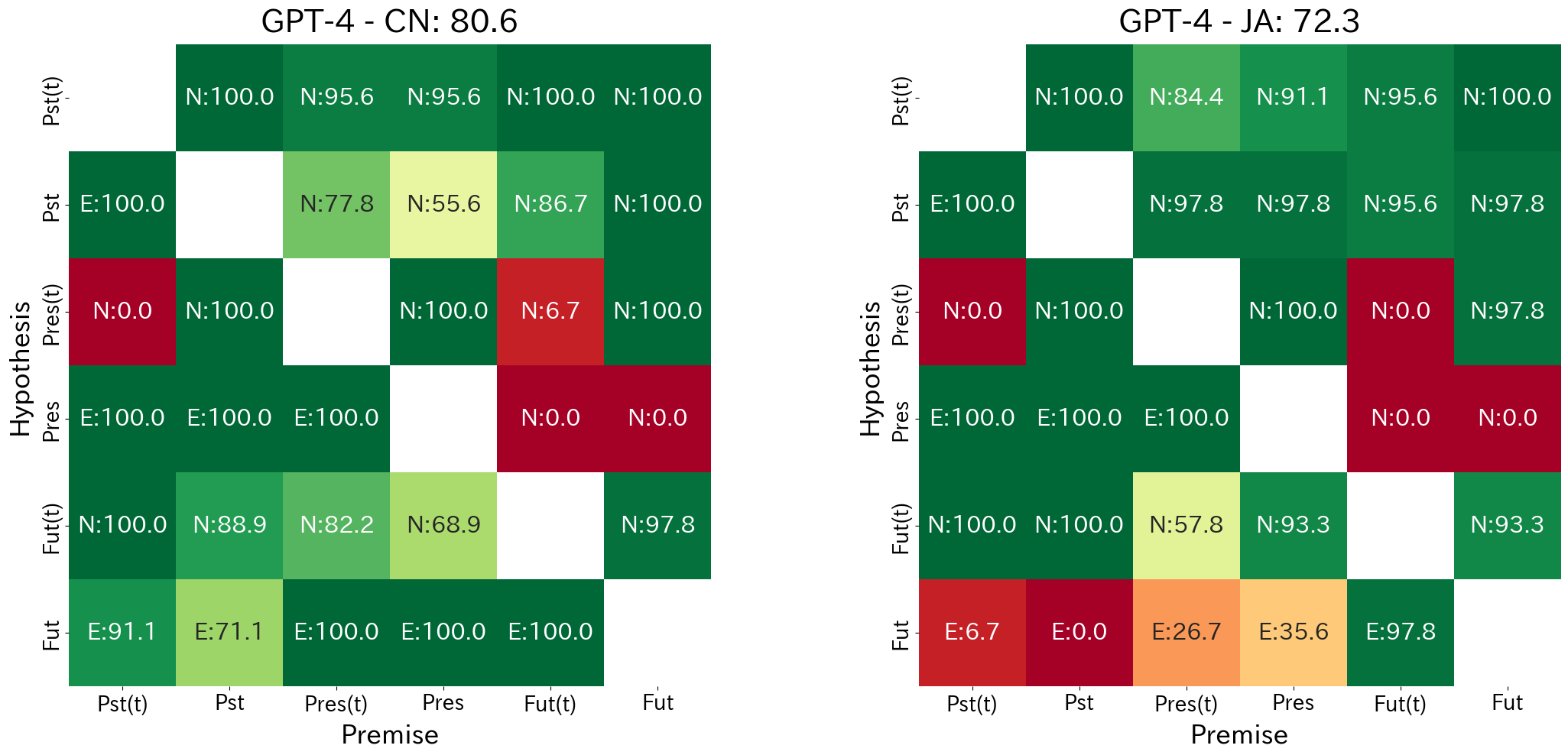}
    \caption{Detailed results from GPT-4 in Chinese and Japanese.The overall accuracy is shown in the title. E/N:number in cells shows the gold label and the accuracy for each $(P,H)$ pair.}
    \label{fig: Results of GPT-4}
\end{figure}

\paragraph{Comparison between models}
As shown in Table~\ref{tab:model_language_accuracy}, Claude 3.5 achieved the best overall performance, outperforming GPT-4—the second-best model—by over 10\% in both Chinese and Japanese.

Most models performed similarly on Chinese and Japanese, with accuracy differing by less than 5\%.
However, Llama-8B was a notable outlier, showing a large performance gap of 26.2\% (Chinese: 37.3\%, Japanese: 65.6\%).
Notably, Llama-8B shows an accuracy gap of nearly 80\% between instances labeled as entailment and those labeled as non-entailment (See Table~\ref{tab:Analyzing Llama8B}).
Given that the contexts in which the perfect aspect appears in Chinese are more homogeneous, this result suggests that multilingual models with smaller parameter sizes may struggle to generalize the meaning of the perfect aspect in Chinese.

Furthermore, LLMs with enhanced monolingual capability (Qwen3 and Swallow) exhibit a negative correlation between accuracy and model size. We aim to explore this phenomenon in greater depth in future studies.

\paragraph{Comparison based on linguistic phenomena}
When the tense of the premise and the hypothesis is the same, models with parameter sizes over 32 billion achieve near-perfect accuracy, while those with lower parameter sizes still struggle with it. 
Example (\ref{exp:(Pst(t), Pst)_CN}) shows a case of ($P$: Pst(t), $H$: Pst).

\begin{exe}
    \ex \label{exp:(Pst(t), Pst)_CN}
    Pst(t): 太郎 上周 回到 家 的\cn{\cn{时候}}，花子\cn{已经}\textbf{死三天了}。\\
    ``When Taro came home last week, Hanako had already been dead for 3 days."\\
    $\not\Rightarrow$ Pst: 太郎 上周 回到 家 的\cn{时候}，花子\cn{已经}\textbf{死了}。 \\
    ``When Taro came home last week, Hanako had already been dead."
\end{exe}

\noindent This demonstrates that models with larger parameter sizes can capture the semantic nuances introduced by temporal adverbs.

However, when the tense of the premise and the hypothesis differ,
the situation becomes more complex.
In cases where the premise is the past or future and the hypothesis is the present (e.g.,  ($P$: Fut, $H$: Pres), we found all models except Claude3.5 consistently predict \textit{entailment} (See Table \ref{tab:model_time_performance_clean}).
One possible reason is that the models rely on lexical overlap heuristics mentioned in \cite{mccoy2019} to solve these problems.
In Chinese, since the aspect marker ``\textit{le}'' applies across all tenses, lexical overlap naturally occurs.
In Japanese, sentence pairs where both the premise and the hypothesis use the same perfect aspect marker (e.g., (Fut, Pres)) involve lexical overlap.
Examples (\ref{exp:(Pst(t), Pres(t))_CN}) and (\ref{exp:(Fut(t), Pres(t))_JP}) illustrate cases where lexical overlap occurs.

\begin{exe}
    \ex \label{exp:(Pst(t), Pres(t))_CN}
    Fut: 太郎明年大学\cn{毕业}的\cn{时候}，花子\cn{已经}辞\cn{职}\textbf{了}。\\
    ``When Taro graduates from college next year, Hanako will have already quit her job."\\
    $\not\Rightarrow$ Pres: 花子\cn{已经}辞\cn{职}\textbf{了}。 \\
    ``Hanako has already quit her job."
\end{exe}

\begin{exe}
    \ex \label{exp:(Fut(t), Pres(t))_JP}
    Fut: 太郎が来年大学を卒業するとき、花子はとっくに会社を辞め\textbf{ている}。\\
    ``When Taro graduates from college next year, Hanako will already have quit her job."\\
    $\not\Rightarrow$ Pres: 花子は会社を辞め\textbf{ている}。\\
    ``Hanako has already quit her job."
\end{exe}

To our surprise, in Japanese cases where the premise and hypothesis use different tense markers, models still tend to incorrectly predict \textit{entailment}, as illustrated by Example (\ref{exp:(Pst(t), Pres(t))_JP}), in which ``\textit{-tei-ta}" is used in the premise and ``\textit{-tei-ru}" in the hypothesis.
This result may suggest that the models' low accuracy in handling the perfect aspect in both Chinese and Japanese is not merely a consequence of heuristic biases, but also reflects their incomplete understanding of the semantic distinction between the Japanese perfect aspect marker ``\textit{-tei-ta}" and the simple past marker  ``\textit{-tei-ru}".

\begin{exe}
    \ex \label{exp:(Pst(t), Pres(t))_JP}
    Pst(t): 太郎が先週に家に帰ったとき、花子は既に三日前に死ん\textbf{でいた}。\\
    ``When Taro came home last week, Hanako had already been dead for 3 days."\\
    $\not\Rightarrow$ Pres(t): 花子は三日前に死ん\textbf{でいる}。 \\
    ``Hanako has already been dead for 3 days."
\end{exe}

\section{Conclusion}
\label{sec:conc}
In this study, we presented a bilingual NLI dataset targeting the interpretation of the perfect aspect with punctual verbs in Japanese and Chinese. 
Our results show that even state-of-the-art LLMs often fail to capture the correct temporal relations, especially when tense and reference times differ between sentences. 
Our findings highlight the need for evaluation benchmarks that are both linguistically diverse and sensitive to temporal inference.

\section{Limitation and Future Work}
\label{sec:limit}
One limitation of this study is that our experiments deliberately include only punctual, irreversible verbs (e.g., die) to avoid truth-conditional ambiguities. 
Consequently, our findings do not yet generalize to verbs that occur in perfect-progressive constructions. Extending coverage to such verb classes is left for future work.

Another limitation is that our experiments are only performed in a zero-shot setting. We plan to expand the range of prompt formats used in future experiments.

Finally, some phenomena highlighted in Section \ref{sec:result} remain speculative, most notably the negative scaling trend observed for the Qwen3 series and the Swallow series. 
We will design additional controlled experiments to validate or refute these hypotheses.

\section{Acknowledgement}
We would like to express our heartfelt gratitude to Koki Shibata, Tomoki Doi, Taiga Someya, Daiki Matsuoka, and Izumi Konishi for their generous support and invaluable assistance in both the discussions and the writing process of this research. Without them, this work would not have
been possible.
We also thank the three anonymous reviewers for their helpful comments and feedback. This work was partially supported by the Institute for AI and Beyond of the University of Tokyo, and JSPS KAKENHI grant number JP24H00809.
\clearpage

\bibliographystyle{acl_natbib}
\bibliography{aclanthology,acl2021}

\clearpage
\appendix
\section{Prompts}
\label{sec:prompts}

Chinese: \\
    \cn{指示: 从 entailment, non-entailment 中回答前提和假设的关系.不需要给出解释.}\\
    \emph{限制：} \\
    \cn{- 如果能够通过逻辑知识或常识性知识从前提推导出假设，则输出 entailment.}\\ 
    \cn{- 如果前提成立无法保证假设成立,则输出 non-entailment.}\\ 
    \cn{- 前提和假设中没有省略任何时间成分.}\\ 
    \cn{- 前提和假设的发话时点为现在.}\\
    前提: \{premise\} \\
    \cn{假设:} \{hypothesis\} \\
    答案: \\
---------------------------------\\
Japanese:\\
    指示: 前提と仮説の関係を entailment,non-entailment の中から回答してください.説明は不要です.\\
    \emph{制約：} \\
    - 前提から仮説が,論理的知識や常識的知識を用いて導出可能である場合は entailment と出力 \\ 
    - 前提が成り立つとしても仮説が必ずしも成り立たない場合は non-entailment と出力 \\ 
    - 前提と仮説には,時間的な成分を省略していない \\ 
    - 前提と仮説の発話時を現在とする \\
    前提: \{premise\} \\
    仮説: \{hypothesis\} \\
    答え: \\
------------------------------------\\
English translation: \\
Instruction: Answer the relationship between the premise and the hypothesis with one of the following: entailment or non-entailment. No explanation is needed.\\
Constraints:\\
- If the hypothesis can be deduced from the premise through logical reasoning or common sense knowledge, output entailment.\\
- If the truth of the premise does not guarantee the truth of the hypothesis, output non-entailment.
- There is no omission of any temporal information in both the premise and hypothesis.\\
- The utterance time for both the premise and hypothesis is the present.\\
Premise: \{premise\}\\
Hypothesis: \{hypothesis\}\\
Answer:\\


\section{Templates}
Table~\ref{tab:question_example} shows all $(P,H)$ templates and their labels in our dataset.
Table~\ref{tab:dataset_example_cn} and Table~\ref{tab:dataset_example_jp} show Chinese and Japanese sentence templates used to create our dataset.
\label{sec:templates}
\renewcommand{\arraystretch}{0.6} 
\begin{table*}[h!]
\footnotesize 
\centering
{\footnotesize
\begin{tabular}{llll}
    \toprule
    \textbf{Premise} & \textbf{Hypothesis} & \textbf{Example} & \textbf{Label} \\
    \midrule
    Pst(t) & & When Taro got his PhD last year, Hanako had already been dead for 3 months. & \\
                      & Pst    & When Taro got his PhD last year, Hanako had already been dead. & Entailment \\
                      & Pres(t) & Hanako has already been dead for 3 months. & Non-Entailment \\
                      & Pres    & Hanako has already been dead. & Entailment \\
                      & Fut(t) &  When Taro gets his PhD next year, Hanako will have already been dead for 3 months.& Non-Entailment \\
                      & Fut    & When Taro gets his PhD next year, Hanako will have already been dead. & Entailment \\
    \addlinespace[0.5em]
    Pst    & & When Taro got his PhD last year, Hanako had already been dead. & \\
                      & Pst(t) & When Taro got his PhD last year, Hanako had already been dead for 3 months. & Non-Entailment \\
                      & Pres(t) & Hanako has already been dead for 3 months. & Non-Entailment \\
                      & Pres    & Hanako has already been dead. & Entailment \\
                      & Fut(t) & When Taro gets his PhD next year, Hanako will have already been dead for 3 months. & Non-Entailment \\
                      & Fut    & When Taro gets his PhD next year, Hanako will have already been dead. & Entailment \\
    \addlinespace[0.5em]
    Pres(t) & & Hanako has already been dead for 3 months.& \\
                      & Pst(t) & When Taro got his PhD last year, Hanako had already been dead for 3 months. & Non-Entailment \\
                      & Pst    & When Taro got his PhD last year, Hanako had already been dead. & Non-Entailment \\
                      & Pres    & Hanako has already been dead. & Entailment \\
                      & Fut(t) & When Taro gets his PhD next year, Hanako will have already been dead for 3 months. & Non-Entailment \\
                      & Fut    & When Taro gets his PhD next year, Hanako will have already been dead. & Entailment \\
    \addlinespace[0.5em]
    Pres    & &Hanako has already been dead. & \\
                      & Pst(t) & When Taro got his PhD last year, Hanako had already been dead for 3 months. & Non-Entailment \\
                      & Pst    & When Taro got his PhD last year, Hanako had already been dead. & Non-Entailment \\
                      & Pres(t) & Hanako has already been dead for 3 months. & Non-Entailment \\
                      & Fut(t) & When Taro gets his PhD next year, Hanako will have already been dead for 3 months. & Non-Entailment \\
                      & Fut    & When Taro gets his PhD next year, Hanako will have already been dead. & Entailment \\
    \addlinespace[0.5em]
    Fut(t) & & When Taro gets his PhD next year, Hanako will have already been dead for 3 months. &\\
                      & Pst(t) & When Taro got his PhD last year, Hanako had already been dead for 3 months. & Non-Entailment \\
                      & Pst    & When Taro got his PhD last year, Hanako had already been dead. & Non-Entailment \\
                      & Pres(t) & Hanako has already been dead for 3 months. & Non-Entailment \\
                      & Pres    & Hanako has already been dead. & Non-Entailment \\
                      & Fut    & When Taro gets his PhD next year, Hanako will have already been dead. & Entailment \\
    \addlinespace[0.5em]
    Fut    & & When Taro gets his PhD next year, Hanako will have already been dead. &\\
                      & Pst(t) & When Taro got his PhD last year, Hanako had already been dead for 3 months. & Non-Entailment \\
                      & Pst    & When Taro got his PhD last year, Hanako had already been dead. & Non-Entailment \\
                      & Pres(t) & Hanako has already been dead for 3 months. & Non-Entailment \\
                      & Pres    & Hanako has already been dead. & Non-Entailment \\
                      & Fut(t) & When Taro gets his PhD next year, Hanako will have already been dead for 3 months. & Non-Entailment \\
    \bottomrule
\end{tabular}
}
\caption{All $(P,H)$ templates and their labels. Here, we only present the English translation of one example to illustrate the correspondence between the $(P,H)$ pair and their label in our dataset. As mentioned in Section~\ref{sec:dataset}, the label remains unchanged even when different punctual verbs are used.}
\label{tab:question_example}
\end{table*}
\begin{table*}[h!]
\centering
\scriptsize
{\scriptsize
\begin{tabular}{p{1cm}|p{6cm}|p{6.5cm}}
    \toprule
    \textbf{Category} & \textbf{Template} & \textbf{Example} \\
    \midrule
    Pst(t) & \textcolor{red}{\texttt{[Event-Past]}}的\cn{时候}，\textcolor{blue}{\texttt{[NP]}}\cn{已经}\textcolor{orange}{\texttt{[VP]}}\textcolor{green}{\texttt{[TIME]}}了 & \textcolor{red}{田中上周搬家}的\cn{时候}，\textcolor{blue}{山本}\cn{已经}\textcolor{orange}{合格大学}\textcolor{green}{一周}了 \\
    
    Pst & \textcolor{red}{\texttt{[Event-Past]}}的\cn{时候}，\textcolor{blue}{\texttt{[NP]}}\cn{已经}\textcolor{orange}{\texttt{[VP]}}了 & \textcolor{red}{田中上周搬家}的\cn{时候}，\textcolor{blue}{山本}\cn{已经}\textcolor{orange}{合格大学}了\\
    
    Pres(t) & \textcolor{blue}{\texttt{[NP]}}\cn{已经}\textcolor{orange}{\texttt{[VP]}}\textcolor{green}{\texttt{[TIME]}}了 & \textcolor{blue}{山本}\cn{已经}\textcolor{orange}{合格大学}\textcolor{green}{一周}了\\
    
    Pres & \textcolor{blue}{\texttt{[NP]}}\cn{已经}\textcolor{orange}{\texttt{[VP]}}了 & \textcolor{blue}{山本}\cn{已经}\textcolor{orange}{合格大学}了\\
    
    Fut(t) & \textcolor{red}{\texttt{[Event-Future]}}的\cn{时候}，\textcolor{blue}{\texttt{[NP]}}\cn{已经}\textcolor{orange}{\texttt{[VP]}}\textcolor{green}{\texttt{[TIME]}}了 & \textcolor{red}{佐藤下个月换工作}的\cn{时候}，\textcolor{blue}{山本}\cn{已经}\textcolor{orange}{合格大学}\textcolor{green}{一周}了\\
    
    Fut  & \textcolor{red}{\texttt{[Event-Future]}}的\cn{时候}，\textcolor{blue}{\texttt{[NP]}}\cn{已经}\textcolor{orange}{\texttt{[VP]}}了 & \textcolor{red}{佐藤下个月换工作}的\cn{时候}，\textcolor{blue}{山本}\cn{已经}\textcolor{orange}{合格大学}了\\
    \bottomrule
\end{tabular}
}
\caption{Sentence Templates for Chinese.}
\label{tab:dataset_example_cn}
\end{table*}

\begin{table*}[h!]
\centering
\scriptsize 
\label{tab:tedataset_examplate_jp_compact}
\begin{tabular}{p{1cm}|p{7cm}|p{7cm}} 
    \toprule
    \textbf{Category} & \textbf{Template} & \textbf{Example} \\
    \midrule
    Pst(t) & \textcolor{red}{\texttt{[Event-Past]}}とき、\textcolor{blue}{\texttt{[NP]}}は\textcolor{teal}{\texttt{[TIME]}}前にすでに\textcolor{orange}{\texttt{[V-teita]}} & \textcolor{red}{田中が先月引っ越した}とき、\textcolor{blue}{山本}は\textcolor{teal}{一週間前}にすでに\textcolor{orange}{大学に合格していた} \\
    
    Pst & \textcolor{red}{\texttt{[Event-Past]}}とき、\textcolor{blue}{\texttt{[NP]}}はすでに\textcolor{orange}{\texttt{[V-teita]}} & \textcolor{red}{田中が先月引っ越した}とき、\textcolor{blue}{山本}はすでに\textcolor{orange}{大学に合格していた} \\
    
    Pres(t) & \textcolor{blue}{\texttt{[NP]}}は\textcolor{teal}{\texttt{[TIME]}}前に\textcolor{orange}{\texttt{[V-teiru]}} & \textcolor{blue}{山本}は\textcolor{teal}{一週間前}に\textcolor{orange}{大学に合格している} \\
    
    Pres & \textcolor{blue}{\texttt{[NP]}}は\textcolor{orange}{\texttt{[V-teiru]}} & \textcolor{blue}{山本}は\textcolor{orange}{大学に合格している} \\
    
    Fut(t) & \textcolor{red}{\texttt{[Event-Future]}}とき、\textcolor{blue}{\texttt{[NP]}}は\textcolor{teal}{\texttt{[TIME]}}前に\textcolor{orange}{\texttt{[V-teiru]}} & \textcolor{red}{佐藤が来月転職する}とき、\textcolor{blue}{山本}は\textcolor{teal}{一週間前}に\textcolor{orange}{大学に合格している} \\
    
    Fut & \textcolor{red}{\texttt{[Event-Future]}}とき、\textcolor{blue}{\texttt{[NP]}}はとっくに\textcolor{orange}{\texttt{[V-teiru]}} & \textcolor{red}{佐藤が来月転職する}とき、\textcolor{blue}{山本}はとっくに\textcolor{orange}{大学に合格している} \\
    \bottomrule
\end{tabular}
\caption{Sentence Templates for Japanese.}
\label{tab:dataset_example_jp}
\end{table*}

\section{Detailed Results}
\begin{figure*}[h!]
    \centering
    \includegraphics[width=1\linewidth]{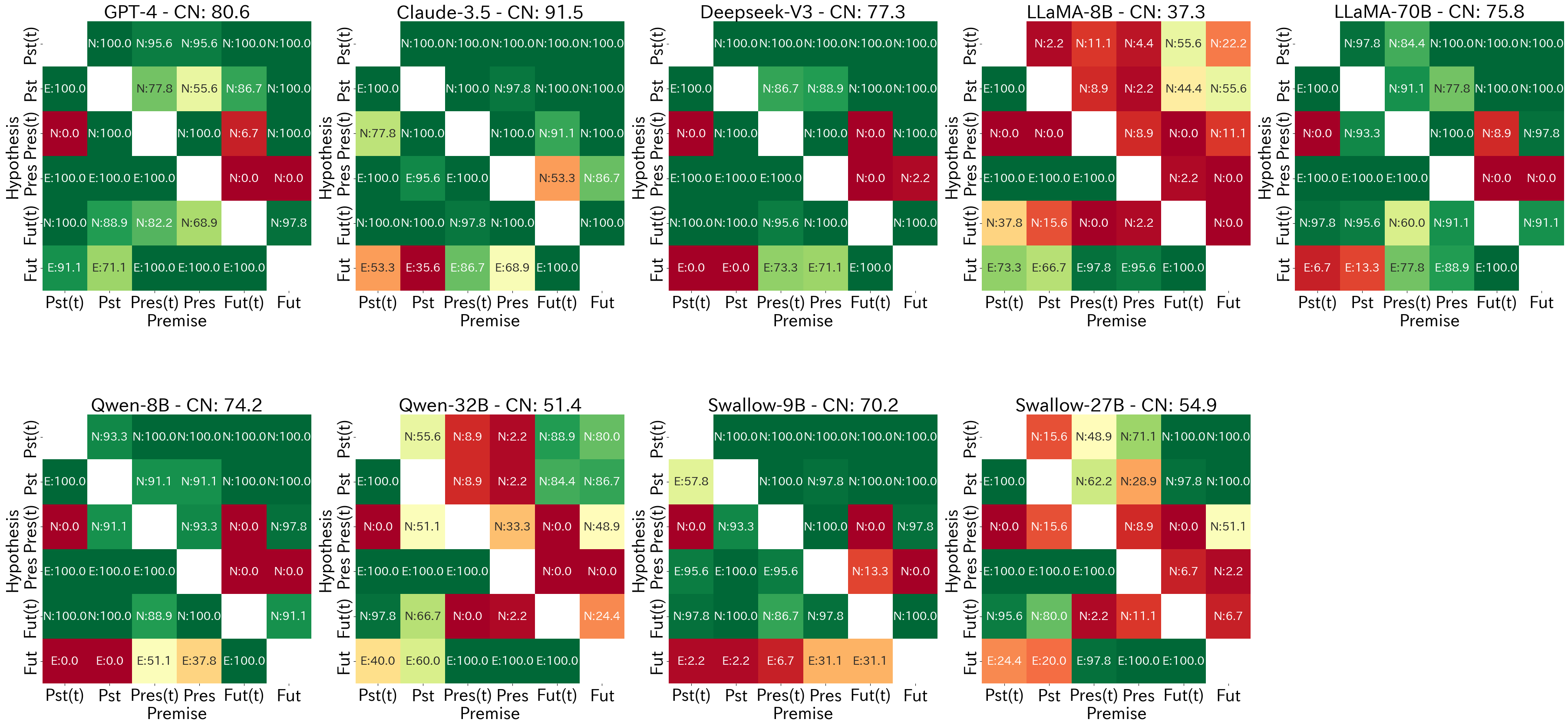}
    \caption{Results on our Chinese dataset. The overall accuracy is shown in the title. E/N:number in cells shows the gold label and the accuracy for each $(P,H)$ pair.}
    \label{fig:results_cn}
    \includegraphics[width=1\linewidth]{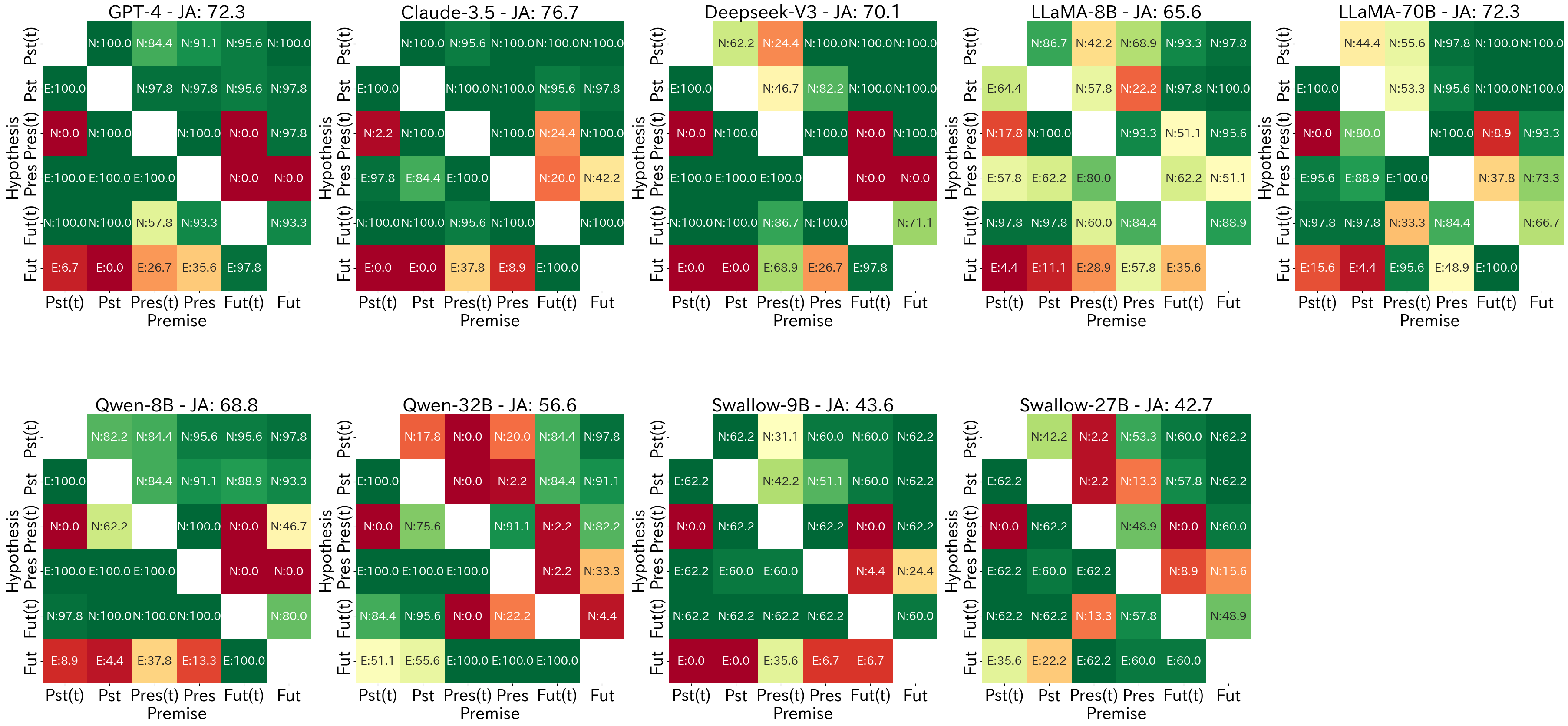}
    \caption{Results on our Japanese dataset.  The overall accuracy is shown in the title. E/N:number in cells shows the gold label and the accuracy for each $(P,H)$ pair.}
    \label{fig:results_jp}
\end{figure*}

Figure~\ref{fig:results_cn} and Figure~\ref{fig:results_jp} show detailed results of all models under our dataset.
\label{app:detailed results of Chinese}

\end{document}